\documentclass[11pt,a4paper]{article}
\usepackage[hyperfootnotes=false]{hyperref}
\usepackage[hyperref]{emnlp-ijcnlp-2019}
\usepackage{times}
\usepackage{latexsym}

\usepackage{url}
\usepackage{amsmath}
\usepackage{bm}
\usepackage{graphicx}
\usepackage{amsfonts}
\usepackage{multirow}

\aclfinalcopy 

\title{Data Augmentation with Atomic Templates for Spoken Language Understanding}

\author{Zijian Zhao\thanks{\ \ Zijian Zhao and Su Zhu are co-first authors and contribute equally to this work.}, Su Zhu\footnotemark[1] \and Kai Yu\thanks{\ \ The corresponding author is Kai Yu.}\\
  MoE Key Lab of Artificial Intelligence\\
  SpeechLab, Department of Computer Science and Engineering\\
  Shanghai Jiao Tong University, Shanghai, China\\
  {\tt \{1248uu,paul2204,kai.yu\}@sjtu.edu.cn} \\}

\date{}

\begin{document}
\maketitle

\begin{abstract}
Spoken Language Understanding (SLU) converts user utterances into structured semantic representations. Data sparsity is one of the main obstacles of SLU due to the high cost of human annotation, especially when domain changes or a new domain comes. In this work, we propose a data augmentation method with atomic templates for SLU, which involves minimum human efforts. The atomic templates produce exemplars for fine-grained constituents of semantic representations. We propose an encoder-decoder model to generate the whole utterance from atomic exemplars. Moreover, the generator could be transferred from source domains to help a new domain which has little data. Experimental results show that our method achieves significant improvements on DSTC 2\&3 dataset which is a domain adaptation setting of SLU. 

\end{abstract}
\section{Introduction}
\label{sec:intro}
The SLU module is a key component of spoken dialogue system, parsing user utterances into corresponding semantic representations in a narrow domain. The typical semantic representation for SLU could be semantic frame~\cite{tur2011spoken} or dialogue act~\cite{young2007cued}. In this paper, we focus on SLU with the dialogue act that a sentence is labelled as a set of \emph{act-slot-value} triples. For example, the utterance \textit{``Not Chinese but I want Thai food please''} has an annotation of \textit{``deny(food=Chinese), inform(food=Thai)''}. 

Deep learning has achieved great success in the SLU field~\cite{mesnil2015using,liu2016attention,zhao2019hierarchical}. However, it is notorious for requiring large labelled data, which limits the scalability of SLU models. Despite recent advancements and tremendous research activity in semi-supervised learning and domain adaptation, the deep SLU models still require massive amounts of labelled data to train.

\begin{figure}[tpb]
\centering
\includegraphics[width=0.75\linewidth]{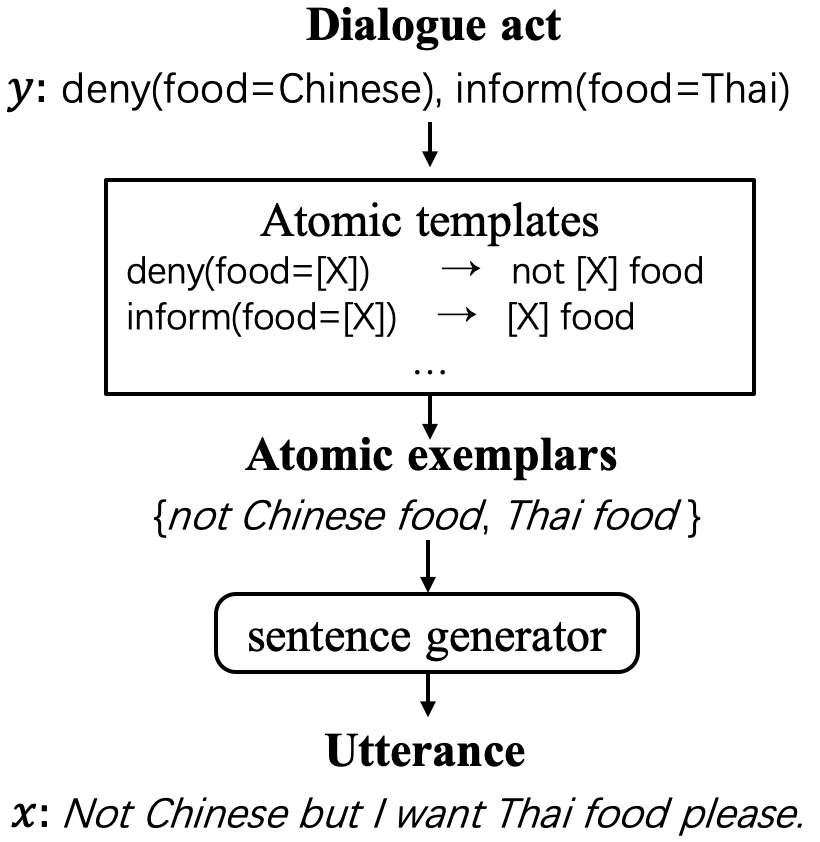}
\caption{Workflow of the data augmentation with atomic templates for SLU.}
\label{fig:workflow}
\end{figure}

Therefore, data augmentation for SLU becomes appealing, and it needs three kinds of capabilities:
\begin{itemize}
    \item \textbf{Expression diversity}: There are always various expressions for the same semantic meaning. Lack of expression diversity is an obstacle of SLU.
    \item \textbf{Semantic diversity}: The data augmentation method should generate data samples with various semantic meanings.
    \item \textbf{Domain adaptation}: The method should be able to utilize data from other domains and adapt itself to a new domain rapidly.
\end{itemize}

In this work, we propose a new data augmentation method for SLU, which consists of atomic templates and a sentence generator (as shown in Figure \ref{fig:workflow}). The method starts from dialogue acts which are well-structured and predefined by domain experts. Rich and varied dialogue acts can be created automatically, which is guided by a domain ontology.

To enhance the capability of domain adaptation for the sentence generator, we propose to first interpret dialogue acts in natural language so that new dialogue acts can be understood by the generator. However, interpreting a dialogue act at sentence level costs a lot, atomic templates are exploited to alleviate human efforts. The atomic templates are created at phrase level, which produce exemplars for fine-grained constituents (i.e. \emph{act-slot-value} triple) of semantic representations (i.e. dialogue act). Thus, the sentence generator can be an encoder-decoder model, paraphrasing a set of atomic exemplars to an utterance.



We evaluate our approach on DSTC 2\&3 dataset~\cite{henderson2014second,henderson2014third}, a benchmark of SLU including domain adaptation setting. The results show that our method can obtain significant improvements over strong baselines with limited training data.

\section{Related Work}

\noindent\textbf{Data augmentation for SLU}\quad \citet{hou2018sequence} use a sequence-to-sequence model to translate between sentence pairs with the same meaning. However, it cannot generate data for new semantic representations. \citet{yoo2018data} use a variational autoencoder to generate labelled language. It still lacks variety of semantic meanings. Let $x$ and $y$ denote an input sentence and the corresponding semantic representation respectively. These two works estimate $p(x, y)$ with generation models to produce more labelled samples, while they have two drawbacks. 1) The models cannot control which kinds of semantic meaning should be generated for supply. 2) The models cannot generate data for new semantic representations which may contain out-of-vocabulary (OOV) labels. We propose to generate utterance based on semantic representations by estimating $p(x|y)$, because $y$ is well-structured and easy to be synthesized. To overcome the OOV problem of semantic labels, we first map them to atomic exemplars with a little of human effort.


\noindent\textbf{Zero-shot learning of SLU}\quad Besides data augmentation, zero-shot learning of SLU \cite{ferreira2015zero,yazdani2015model}  is also related, which can adapt to unseen semantic labels. \citet{yazdani2015model} exploit a binary classifier for each possible \emph{act-slot-value} triple to predict whether it exists in the input sentence. \citet{zhao2019hierarchical} propose a hierarchical decoding model for SLU. However, they still have problem with new \emph{act} and \emph{slot}. \citet{Bapna2017towards,lee2018zero,sz128-zhu-sigdial18} try to solve it with textual slot descriptions. In this paper, we propose atomic templates to describe \emph{act-slot-value} triples but not separate slots or acts.
\section{Method}

In this section, our data augmentation method with atomic templates will be introduced, which generates additional data samples for training SLU model.

\subsection{SLU Model}

For dialogue act based SLU, we adopt the hierarchical decoding (HD) model~\cite{zhao2019hierarchical} which dynamically parses \emph{act}, \emph{slot} and \emph{value} in a structured way. The model consists of four parts:
\begin{itemize}
    \item A shared utterance encoder, bidirectional LSTM recurrent neural network (BLSTM) \cite{Graves2012Supervised};
    \item An \emph{act} type classifier on the utterance;
    \item A \emph{slot} type classifier with the utterance and an act type as inputs;
    \item A \emph{value} decoder with the utterance and an \emph{act-slot} type pair as inputs.
\end{itemize}
The \emph{value} decoder generates word sequence of the value by utilizing a sequence-to-sequence  model with attention \cite{luong2015effective} and pointer network \cite{vinyals2015pointer} which helps handling out-of-vocabulary (OOV) values.

\subsection{Data Generation with Atomic Templates}

As illustrated in Figure~\ref{fig:workflow}, the workflow of the data augmentation is broken down into: 1) mapping \emph{act-slot-value} triples to exemplars with atomic templates, and 2) generating the corresponding utterance depending on the atomic exemplars. 

Let $x=x_1 \cdots x_{|x|}$ denote the utterance (word sequence), and $y=\{y_1, \cdots, y_{|y|}\}$ denote the dialogue act (a set of \emph{act-slot-value} triples). We wish to estimate $p(x|y)$, the conditional probability of utterance $x$ given dialogue act $y$.


However, there are also some disadvantages to directly using dialogue acts as inputs:
\begin{itemize}
\item The dialogue acts from different narrow domains may conflict, e.g. different slot names for the same meaning.
\item Act and slot types may not be defined in a literal way, e.g. using arbitrary symbols like ``city\_1'' and ``city\_2'' to represent city names in different contexts.
\end{itemize}
Thus, it is hard to adapt the model $p(x|y)$ to new act types, slot types, and domains.

We propose to interpret the dialogue act in short natural language, and then rephrase it to the corresponding user utterance. While interpreting the dialogue act at sentence level costs as much as building a rule-based SLU system, we choose to interpret  \emph{act-slot-value} triples with atomic templates which involve minimum human efforts.

\subsubsection{Atomic templates}

\begin{table}[tpb]
\centering
\small
\begin{tabular}{|c|c|}
\hline
\textbf{Triple}             & \textbf{Templates}  \\
\hline
bye()               & \begin{tabular}[c]{@{}c@{}}goodbye\\  bye\end{tabular} \\
\hline
request(addr)       & \begin{tabular}[c]{@{}c@{}}the address\\ what's the address\end{tabular} \\ 
\hline
\text{inform(food=[food])} & \begin{tabular}[c]{@{}c@{}} \text{[food]}\\ \text{[food] food} \end{tabular} \\
\hline
inform(hastv=true)  & \begin{tabular}[c]{@{}c@{}}television\\ with a television\end{tabular}                     \\
\hline
\end{tabular}
\caption{Examples of atomic templates in DSTC 2\&3 dataset. ``[food]'' is an arbitrary value of slot ``food''.}
\label{tab:atomic_template}
\end{table}

Table \ref{tab:atomic_template} gives some examples of atomic templates used in DSTC 2\&3 dataset. Atomic templates produce a simple description (atomic exemplar $e_i$) in natural language for each \emph{act-slot-value} triple\footnote{It should be noted that an \emph{act-slot-value} triple may have an empty value, e.g. ``request(phone)'' refers to asking for phone number. The triple may also need no slot, e.g. ``bye()''.} $y_i$. If there are multiple templates for triple $y_i$, which generate multiple atomic exemplars $E(y_i)$, we choose the most similar one $e_i=\text{argmax}_{e\in E(y_i)} sim(e, x)$ in the training stage, and randomly select one $e_i$ from $E(y_i)$ in the data augmentation stage. The similarity function $sim(e,x)$ we used is Ratcliff-Obershelp algorithm~\cite{Ratcliff-Obershelp}. Therefore,
$$
p(x|y)=p(x|\{y_1 \cdots y_{|y|}\})=p(x|\{e_1 \cdots e_{|y|}\})
$$

\subsubsection{Sentence generator}

An encoder-decoder model is exploited to generate the utterance based on the set of atomic exemplars by estimating $p(x|\{e_1 \cdots e_{|y|}\})$.

As the set of atomic exemplars is unordered, we encode them independently. For each atomic exemplar $e_i=w_{i1} \cdots w_{iT_i}$ (a sequence of words with length $T_i$), we use a BLSTM to encode it. The hidden vectors are recursively computed at the $j$-th time step via:
\begin{align*}
\overrightarrow{\textbf{h}_{ij}}=&\text{f}_\text{{LSTM}}(\psi(w_{ij}), \overrightarrow{\textbf{h}}_{ij-1}), j=1,\cdots,|T_i|\\
\overleftarrow{\textbf{h}_{ij}}=&\text{b}_\text{{LSTM}}(\psi(w_{ij}), \overleftarrow{\textbf{h}}_{ij+1}), j=|T_i|,\cdots,1\\
\textbf{h}_{ij}=&[\overrightarrow{\textbf{h}}_{ij};\overleftarrow{\textbf{h}}_{ij}]
\end{align*}
where $[\cdot;\cdot]$ denotes vector concatenation, $\psi(\cdot)$ is a word embedding function, $\text{f}_\text{{LSTM}}$ is the forward LSTM function and $\text{b}_\text{{LSTM}}$ is the backward one. The final hidden vectors of the forward and backward passes are utilized to represent each atomic exemplar $e_i$, i.e. $\textbf{c}_{i}=[\overrightarrow{\textbf{h}}_{iT_i};\overleftarrow{\textbf{h}}_{i1}]$.

After encoding all the atomic exemplars, we have a list of hidden vectors $H(y)=[\textbf{h}_{11}\cdots \textbf{h}_{1T_1};\cdots;\textbf{h}_{|y|1}\cdots \textbf{h}_{|y|T_{|y|}}]$. A LSTM model serves as the decoder~\cite{vinyals2015grammar}, generating the utterance $x$ word-by-word:
$$
p(x|y)=p(x|H(y))=\prod_{t=1}^{|x|}p(x_t|x_{<t},H(y))
$$
Before the generation, the hidden vector of the decoder is initialized as $1/|y|\sum_{i=1}^{|y|}\textbf{c}_{i}$, the mean of representations of all the atomic exemplars. The pointer softmax~\cite{gulcehre-EtAl:2016:P16-1} enhanced by a trick of targeted feature dropout ~\cite{xu2018end} is adopted to tackle OOV words, which will switch between generation and copying from the input source dynamically.

\section{Experiment}

\subsection{Data}

In our experiments, we use the dataset provided for the second and third Dialog State Tracking Challenge (DSTC 2\&3)~\cite{henderson2014second,henderson2014third}. DSTC2 contains a large number of training dialogues ($\sim$15.6k utterances) related to restaurant search while DSTC3 is designed to address the problem of adaptation to a new domain (tourist information) with only a small amount of seed data (\emph{dstc3\_seed}, 109 utterances). The manual transcriptions are used as user utterances to eliminate the impact of speech recognition errors.  


We follow the data partitioning policy as \citet{zhu2014semantic}, which randomly selects one-half of DSTC3 test data as the oracle training set ($\sim$9.4k utterances) and leaves the other half as the evaluation set ($\sim$9.2k utterances).

\subsection{Experimental Setup}

Delexicalized triples are used in case of non-enumerable slot, like ``inform(food=[food])'' in Table \ref{tab:atomic_template}. There are 41 and 35 delexicalized triples for DSTC2 and DSTC3 respectively. For each triple, we prepare two short templates on average. Note that, compared to designing sentence-level templates, writing atomic templates for triples at phrase level requires much less human efforts.

For data augmentation of DSTC3, there are two ways to collect extra dialogue acts as inputs\footnote{The data splits, atomic templates and extra dialogue acts will be available at \url{https://github.com/sz128/DAAT_SLU}}:
\begin{itemize}
    \item \textbf{Seed abridgement}: Abridging the delexicalized dialogue acts existing in \emph{dstc3\_seed} data, e.g. a dialogue act with 3 triples \{A,B,C\} can be abridged to obtain \{A,B\}, \{B,C\}, \{A,C\}, \{A\}, \{B\}, \{C\} and itself.
    \item \textbf{Combination}: As all possible triples are predefined in the domain ontology of DSTC3 by experts. We use a general policy of triples combination, which randomly selects at most $N_c$ triples to make up a dialogue act. ($N_c$ is set as $3$ empirically.)
\end{itemize}
Then we fill up each non-enumerable slot in the dialogue act by randomly choosing a value of this slot, which ends when each value appears at least $N_v$ times in all the collected dialogue acts. $N_v$ is set as $3$ empirically. After that, we have 1420 and 20670 dialogue acts from the seed abridgement and combination respectively. New data samples are generated starting from these dialogue acts, through the atomic templates and the sentence generator (e.g. 1-best output is kept).

The SLU model and sentence generator use the same hyper-parameters. The dimension of word embeddings (\textit{Glove}\footnote{\url{http://nlp.stanford.edu/data/glove.6B.zip}} word vectors are used for initialization) is 100 and the number of hidden units is 128. Dropout rate is 0.5 and batch size is 20. Maximum norm for gradient clipping is set to 5 and Adam optimizer is used with an initial learning rate of 0.001. All training consists of 50 epochs with early stopping on the development set. We report F1-score of extracting \emph{act-slot-value} triples by the official scoring script from \url{http://camdial.org/~mh521/dstc/}. 

\subsection{Systems}
We first compare two SLU models to answer why the HD model is chosen for dialogue act based SLU:
\begin{itemize}
    \item \textbf{ZS}: zero-shot learning of SLU \cite{yazdani2015model} which can adapt to unseen dialogue acts.
    \item \textbf{HD}: the hierarchical decoding model~\cite{zhao2019hierarchical} is adopted in our system.
\end{itemize}

We make comparisons of other data augmentation methods and the atomic templates (AT):
\begin{itemize}
    \item \textbf{Naive}: We replace the value simultaneously existing in an utterance and its semantic labels of \emph{dstc3\_seed} by randomly selecting a value of the corresponding slot. It ends when each value appears at least $N_v$ times.
    \item \textbf{Human}: \citet{zhu2014semantic} proposed to design a large number of sentence-level templates for DSTC3 with lots of human efforts.
    \item \textbf{Oracle}: The oracle training set is used which simulates the perfect data augmentation.
\end{itemize}

Without data augmentation, the SLU models are pre-trained on DSTC2 dataset (source domain) and finetuned with \emph{dstc3\_seed} set. In our data augmentation method, the sentence generator based on atomic concepts is also pre-trained on DSTC2 dataset and finetuned with the \emph{dstc3\_seed}. The SLU model is first pre-trained on DSTC2 dataset, then finetuned with the augmented dataset and finally finetuned with the \emph{dstc3\_seed}.

\subsection{Results and Analysis}

\begin{table}[tpb]
\centering
\small
\begin{tabular}{c|c||c}
\hline 
\textbf{SLU}   & \textbf{Augmentation}   & \textbf{F1-score} \\ \hline \hline
ZS & w/o     & 68.3  \\ \hline
\multirow{7}*{HD} & w/o     & 78.5  \\ 
 & Naive  & 82.9 \\ 
 \cline{2-3}
 & AT (seed abridgement)   & 85.5  \\ 
 & AT  (combination)   & 87.9  \\ 
 & AT  (seed abr. + comb.)   &  \textbf{88.6} \\ 
  \cline{2-3} 
 & Human \cite{zhu2014semantic}    & 90.4   \\ 
 & Oracle & 96.9 \\ \hline
\end{tabular}
\caption{SLU performances of different systems on the DSTC3 evaluation set.}
\label{tab:main_results}
\end{table}

The main results are illustrated in Table \ref{tab:main_results}. We can see that: 
\begin{enumerate}
    \item The hierarchical decoding (HD) model gets better performance than the zero-shot learning (ZS) method of SLU.
    \item The seed data $dstc3\_seed$ limits the power of the SLU model, and even the naive augmentation can enhance it.
    \item Our data augmentation method with atomic templates (AT) improves the SLU performance dramatically. One reason may be the generated data has higher variety of semantic meaning than the naive augmentation. \emph{Combination} can make up more dialogue acts and shows better result than \emph{Seed abridgement}, while \emph{Seed abridgement} provides more realistic dialogue acts. Thus, their union gives the best result.
    \item The best performance of our method is close to the human-designed sentence-level templates \cite{zhu2014semantic}, while our approach needs much less human efforts.
\end{enumerate}


\noindent\textbf{Ablation Study}\quad We conduct several ablation studies to analyze the effectiveness of different components of our method, as shown in Table \ref{tab:ablation_study}. By removing SLU model pretraining on DSTC2 (``\emph{- dstc2}'') and finetuning on the seed data (``\emph{- dstc3\_seed}''), we can see a significant decrease in SLU performance. When we subsequently cast aside the sentence generator (``\emph{- sentence generator}'', i.e. using the atomic exemplars as inputs of SLU model directly), the SLU performance decreases by 10.3\%. This shows that the sentence generator can produce more natural utterances. If we replace the atomic exemplars as the corresponding \emph{act-slot-value} triples (``\emph{- atomic templates}''), the SLU performance drops sharply. The reason may be that the atomic templates provide a better description of corresponding semantic meanings than the surfaces of the triples. Examples of generated data samples are in Appendix \ref{app:examples_of_generated_data}.


\begin{table}[tpb]
\centering
\small
\begin{tabular}{|l||c|}
\hline
\textbf{Method}        & \textbf{F1-score} \\ \hline \hline
HD + AT(seed abr. + comb.)   &  88.6 \\ \hline
\ - dstc2   &  86.2 \\ \hline
\ - dstc3\_seed   &  84.5 \\ \hline
\ - dstc2, - dstc3\_seed   &  84.3 \\ \hline 
\ \ \ \ - sentence generator &  74.0 \\ \hline
\ \ \ \ - atomic templates     & 70.5  \\ \hline
\end{tabular}
\caption{SLU performances on the DSTC3 evaluation set when removing different modules of our method.}
\label{tab:ablation_study}
\end{table}

\begin{figure}[]
\centering
\includegraphics[width=0.85\linewidth]{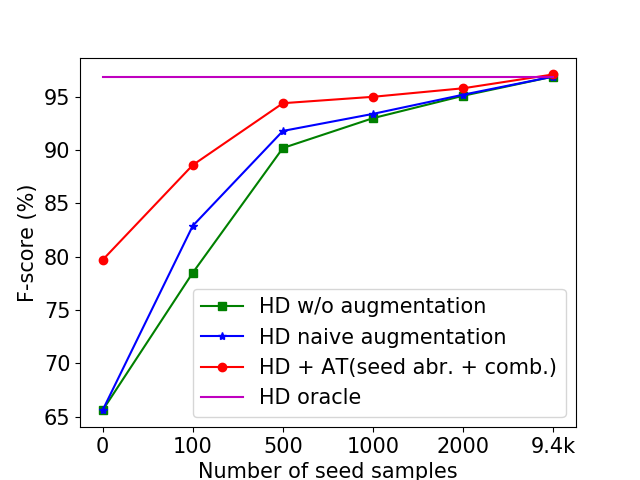}
\caption{SLU performance of different methods with varying number of seed samples randomly selected from the oracle DSTC3 training set.}
\label{fig:number_of_seed_samples}
\end{figure}

\noindent\textbf{Number of seed samples}\quad Figure \ref{fig:number_of_seed_samples} shows the number of seed samples used versus SLU performance on DSTC3 evaluation set. For zero-shot case (no seed samples), our method is much better than the baseline. When the number of seed samples increases, our method outperforms the baseline constantly.
\section{Conclusion}



In this paper, we propose a new data augmentation method with atomic templates for SLU. The atomic templates provide exemplars in natural language for \emph{act-slot-value} triples, involving minimum human efforts. An encoder-decoder model is exploited to generate utterances based on the atomic exemplars. We believe our method can also be applied to SLU systems with other semantic representations (e.g. semantic frame). Experimental results show that our method achieves significant improvements on DSTC 2\&3 dataset, and it is is very effective for SLU domain adaptation with limited data.

\section*{Acknowledgments}
This work has been supported by the National Key Research and Development Program of China (Grant No. 2017YFB1002102) and the China NSFC projects (No. 61573241). We thank the anonymous reviewers for their thoughtful comments and efforts towards improving this manuscript.

\bibliography{ref}
\bibliographystyle{acl_natbib}

\clearpage
\appendix


\section{Examples of generated data}
\label{app:examples_of_generated_data}
\begin{table}[h]
\begin{minipage}{\textwidth}
\begin{tabular}{|l|l|p{6.5cm}|}
\hline
\textbf{Dialogue act}                                                                                                 & \textbf{Atomic exemplars}                                                                                   & \textbf{Generated utterance}                                        \\ \hline
\begin{tabular}[c]{@{}l@{}}thankyou\\ bye\end{tabular}                                                 & \begin{tabular}[c]{@{}l@{}}thank you\\ good bye\end{tabular}                               & Thank you good bye.                               \\ \hline
\begin{tabular}[c]{@{}l@{}}request(addr)\\ request(phone)\end{tabular}                                 & \begin{tabular}[c]{@{}l@{}}the address\\  the phone number\end{tabular}                    & The address and phone number.                      \\ \hline
\begin{tabular}[c]{@{}l@{}}request(addr)\\ request(phone)\end{tabular}                                 & \begin{tabular}[c]{@{}l@{}}whats the address\\ what is the phone number\end{tabular}       & What is the address and phone number?              \\ \hline
\begin{tabular}[c]{@{}l@{}}inform(type=coffeeshop)\\ inform(hasinternet=true)\end{tabular}             & \begin{tabular}[c]{@{}l@{}}cafe\\ has internet connection\end{tabular}                     & I'm looking for a cafe has the internet connection. \\ \hline
\begin{tabular}[c]{@{}l@{}}affirm\\ inform(type=restaurant)\\ inform(pricerange=moderate)\end{tabular} & \begin{tabular}[c]{@{}l@{}}yes\\ looking for a restaurant\\ moderately priced\end{tabular} & Yes, I'm looking for a moderately priced restaurant. \\ \hline
request(childrenallowed)                                                                                & does it allow children                                                                     & Does it {\color{red} have} children?                             \\ \hline
\begin{tabular}[c]{@{}l@{}}request(hastv)\\ request(addr)\end{tabular}                                 & \begin{tabular}[c]{@{}l@{}}does it has a television\\ the address\end{tabular}             & Does it have the television {\color{red} address} and {\color{red} address}?   \\ \hline
\end{tabular}
\caption{Examples of generated data samples for DSTC3. It shows that the generated utterances are associated with the atomic exemplars. For some frequent triples, well-formed utterances can be produced. We can find two bad cases at the end of the table. The reason may be that both ``request(childrenallowed)'' and ``request(hastv);request(addr)'' never appear in the \emph{dstc3\_seed}, thus the sentence generator doesn't fit well.}
\end{minipage}
\end{table}

\end{document}